\documentclass[10pt,twocolumn,letterpaper]{article}

\usepackage{iccv}
\usepackage{times}
\usepackage{epsfig}
\usepackage{graphicx}
\usepackage{amsmath}
\usepackage{amssymb}
\usepackage{algorithm,algpseudocode}
\usepackage{array}
\usepackage{color}
\usepackage{subfiles}
\usepackage{setspace}
\usepackage{soul} 
\usepackage{authblk}

\usepackage[square,numbers]{natbib}
\usepackage[pagebackref=true,breaklinks=true,letterpaper=true,colorlinks,bookmarks=false]{hyperref}

\newcolumntype{C}[1]{>{\centering\let\newline\\\arraybackslash\hspace{0pt}}m{#1}}
\newcolumntype{L}[1]{>{\raggedright\let\newline\\\arraybackslash\hspace{0pt}}m{#1}}
\newcolumntype{R}[1]{>{\raggedleft\let\newline\\\arraybackslash\hspace{0pt}}m{#1}}
\iccvfinalcopy 

\def\iccvPaperID{690} 

\ificcvfinal\fi

\newif\ifpreview

\newcommand{\rvs}[3][N]{\ifpreview#3\else\ifthenelse{\equal{#1}{Y}}{{\color{blue}#3}}{\st{#2}{\color{red} #3}}\fi}
\newcommand{\att}[1]{{\color{black}#1}}

\newcommand{\unsim}{\mathord{\sim}}

\graphicspath{{figure/}{../figure/}}

\begin{document}

\makeatletter
\renewcommand*{\@fnsymbol}[1]{\dag}
\makeatother

\title{Soft Proposal Networks for Weakly Supervised Object Localization}

\author[1]{Yi Zhu}
\author[1]{Yanzhao Zhou}
\author[1]{Qixiang Ye\thanks{Corresponding Authors}}
\author[2]{Qiang Qiu}
\author[1]{Jianbin Jiao$^\dag$}

\affil[1]{University of Chinese Academy of Sciences}
\affil[2]{Duke University
\authorcr\small \{zhuyi215, zhouyanzhao215\}@mails.ucas.ac.cn, \{qxye, jiaojb\}@ucas.ac.cn, qiang.qiu@duke.edu}

\renewcommand\Authands{ and }

\maketitle

\begin{abstract}
Weakly supervised object localization remains challenging, where only image labels instead of bounding boxes are available during training. 
Object proposal is an effective component in localization, but often computationally expensive and incapable of joint optimization with some of the remaining modules. 
In this paper, to the best of our knowledge, we for the first time integrate weakly supervised object proposal into convolutional neural networks (CNNs) in an end-to-end learning manner.  
We design a network component, Soft Proposal (SP), to be plugged into any standard convolutional architecture to introduce the nearly cost-free object proposal, orders of magnitude faster than state-of-the-art methods.
In the SP-augmented CNNs, referred to as Soft Proposal Networks (SPNs), iteratively evolved object proposals are generated based on the deep feature maps then projected back, and further jointly optimized with network parameters, with image-level supervision only. 
Through the unified learning process, SPNs learn better object-centric filters, discover more discriminative visual evidence, and suppress background interference, significantly boosting both weakly supervised object localization and classification performance. 
We report the best results on popular benchmarks, including PASCAL VOC, MS COCO, and ImageNet.
\footnote{Source code is publicly available at \href{http://yzhou.work/SPN}{yzhou.work/SPN}}
\end{abstract}

\section{Introduction}
\begin{figure}[t!]
\begin{center}
\includegraphics[width=0.9\linewidth]{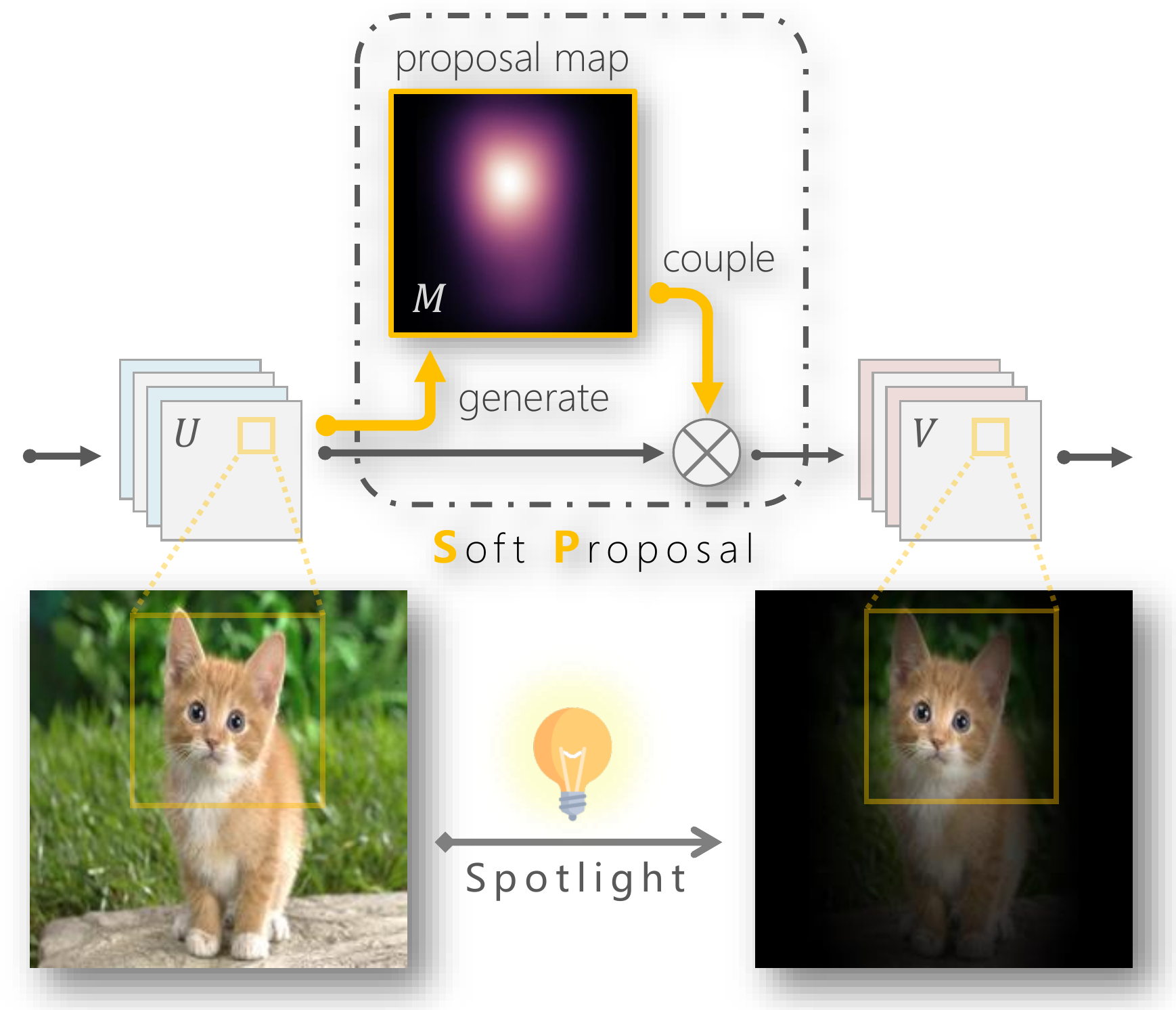}
\end{center}
  \caption{Soft Proposal (SP) module can be inserted after any CNN layer. A proposal map $M$ is generated based on deep feature maps $U$ and then projected back, which results in feature maps $V$. During the end-to-end learning procedure, $M$ iteratively evolves and jointly optimizes with the feature maps to spotlight informative object regions.}
\label{fig:intro-SP}
\vspace{-1em}
\end{figure}

The success of object proposal methods greatly drives the progress of the object localization. With the popularity of deep learning, object detection is evolving from pipelined frameworks \cite{girshick2015fast, girshick2014rich} to unified frameworks \cite{liu2016ssd, redmon2016you, ren2015faster},  
thanks to the unprecedentedly learning capability of convolutional neural networks (CNNs) and abundant object bounding box annotations.

Despite the unified frameworks achieve remarkable performance in supervised object detection, they can not be directly applied to weakly supervised object localization where only image-level labels, \ie, the presence or absence of object categories, are available during training. 

\begin{figure}[t!]
\begin{center}
    \includegraphics[width=0.93\linewidth]{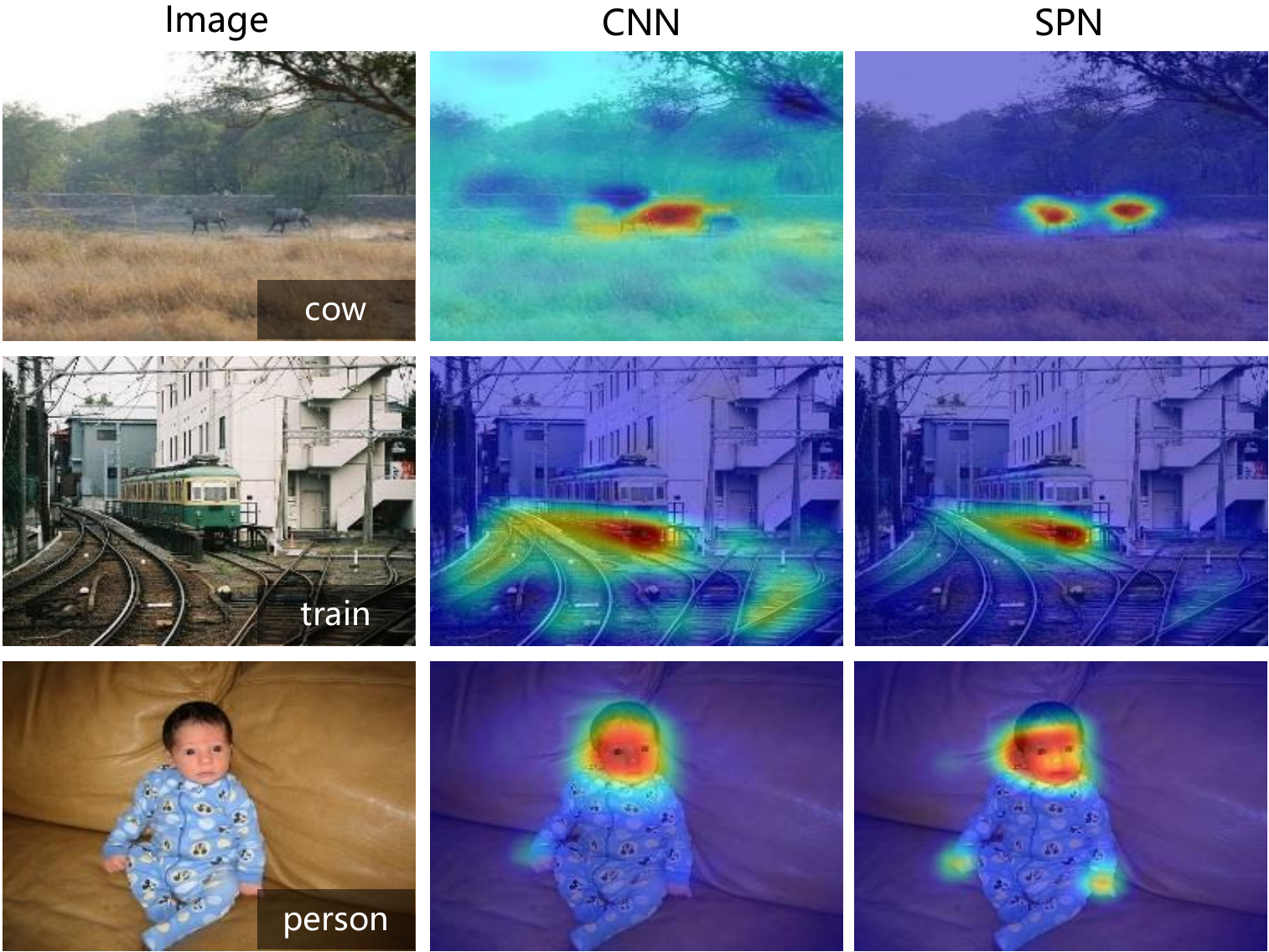}
\end{center}
  \caption{Visualization of Class Activation Maps (CAM) \cite{zhou2015cnnlocalization} for generic CNN and the proposed SPN. CNNs can be misled by noisy backgrounds, \eg, grass for ``cow'', and co-occurrence patterns, \eg, rail for ``train'', and thus miss informative object evidence. In contrast, SPNs focus on informative object regions during training to discover more fine-detailed evidence, \eg, hands for ``person'', while suppressing background interference. Best viewed in color.
  }
\label{fig:intro-examples}
\vspace{-1em}
\end{figure}

To tackle the problem of weakly supervised object localization, many of the conventional methods follow a multi-instance learning (MIL) framework by using object proposal methods \cite{bilen2016weakly, cinbis2017weakly, kantorov2016contextlocnet, wang2014weakly, zhang2015self}. The learning objective is designed to choose an instance (a proposal) from each bag (an image with multiple proposals) to minimize the image classification error; however, the pipelined proposal-and-classification method is sub-optimal as the two steps can not be jointly optimized.
Recent research \cite{bolei2015object} demonstrates that the convolutional filters in CNN can be seen as object detectors and their feature maps can be aggregated to produce Class Activation Map (CAM) \cite{zhou2015cnnlocalization}, which specifies the spatial distribution of discriminative patterns for different image classes. This end-to-end network demonstrates a surprising capability to localize objects under weak supervision. However, without the prior knowledge of informative object regions during training, conventional CNNs can be misled by co-occurrence patterns and noisy backgrounds, Fig.~\ref{fig:intro-examples}. The weakly supervised setting increases the importance of high-quality object proposals, but the problem to integrate the proposal functionality into a unified framework for weakly supervised object localization remains open.

In this paper, we design a network component, Soft Proposal (SP), to be plugged into standard convolutional architectures for nearly cost-free object proposal ($\unsim$0.9ms per image, $10\times$faster than RPN \cite{ren2015faster}, $200\times$faster than EdgeBoxes \cite{zitnick2014edge}), Fig.~\ref{fig:intro-SP}. 
CNNs using SP module are referred to as Soft Proposal Networks (SPNs). In SPNs, iteratively evolved object proposals are projected back on the deep feature maps, and further jointly optimized with network parameters, using image-level labels only.
We further apply the SP module to successful CNNs including CNN-S, VGG, and GoogLeNet, and upgrade them to Soft Proposal Networks (SPNs), which can learn better object-centric filters and discover more discriminative visual evidence for weakly supervised localization tasks. 

The meaning of the word ``soft'' is threefold. First of all, instead of extracting multiple materialized proposal boxes, we predict objectness score for each receptive field, based on the deep feature maps. Next, the proposal couples with deep activation in a probabilistic manner, which not only avoids threshold tuning but also aggregates all information to improve performance. Last but not least, the proposal iteratively evolves along with CNN filters updating.

To summarize, the main contributions of this paper are:
\begin{itemize}
    \item We design a network component, Soft Proposal (SP), to upgrade conventional CNNs to Soft Proposal Networks (SPNs), in which the network parameters can be jointly optimized with the nearly cost-free object proposal.
    \item We upgrade successful CNNs to SPNs, including CNN-S, VGG16, and GoogLeNet, and improve the state-of-the-art of weakly supervised object localization by a significant margin.
\end{itemize}

\section{Related Work}
Weakly supervised object localization problems are often solved with a pipelined approach, \ie, an object proposal method \cite{uijlings2013selective, zitnick2014edge} is first applied to decompose images into object proposals, with which a latent variable learning method, \eg, multi-instance learning (MIL), is used to iteratively perform proposal selection and classifier estimation \cite{cinbis2017weakly, kumar2010self, ye2016self, song2014learning, bilen2014weakly, bilen2016weakly, kantorov2016contextlocnet}. 
With the popularity of deep learning, the pipelined approaches have been evolving to end-to-end MIL networks \cite{oquab2015object, sun2016pronet} by learning convolutional filters as detectors and using response maps to localize objects.

\subsection{Object Proposal}
Conventional object proposal methods, \eg, Selective Search (SS) \cite{uijlings2013selective} and EdgeBoxes (EB) \cite{zitnick2014edge}, use redundant proposals generated with hand-craft features to hypothesize objects locations. 
Region Proposal Network (RPN) regresses object locations using deep convolutional features \cite{ren2015faster}, reports the state-of-the-art proposal performance. 
The success of RPN roots in the localization capability of deep convolutional features; however, such capability is not available until the network is well trained with precise annotations about object locations, \ie, bounding boxes, which limits its applicability to weakly supervised methods.

Our SPN is specified for weakly supervised object localization task with only image-level annotations, \ie, presence or absence of object categories. The key difference between our method to existing ones is that the ``soft'' proposal is an objectness confidence map instead of materialized boxes. Such a proposal couples with convolutional activation and evolves with the deep feature learning.

\begin{figure*}[t!]
\begin{center}
\includegraphics[width=0.88\linewidth]{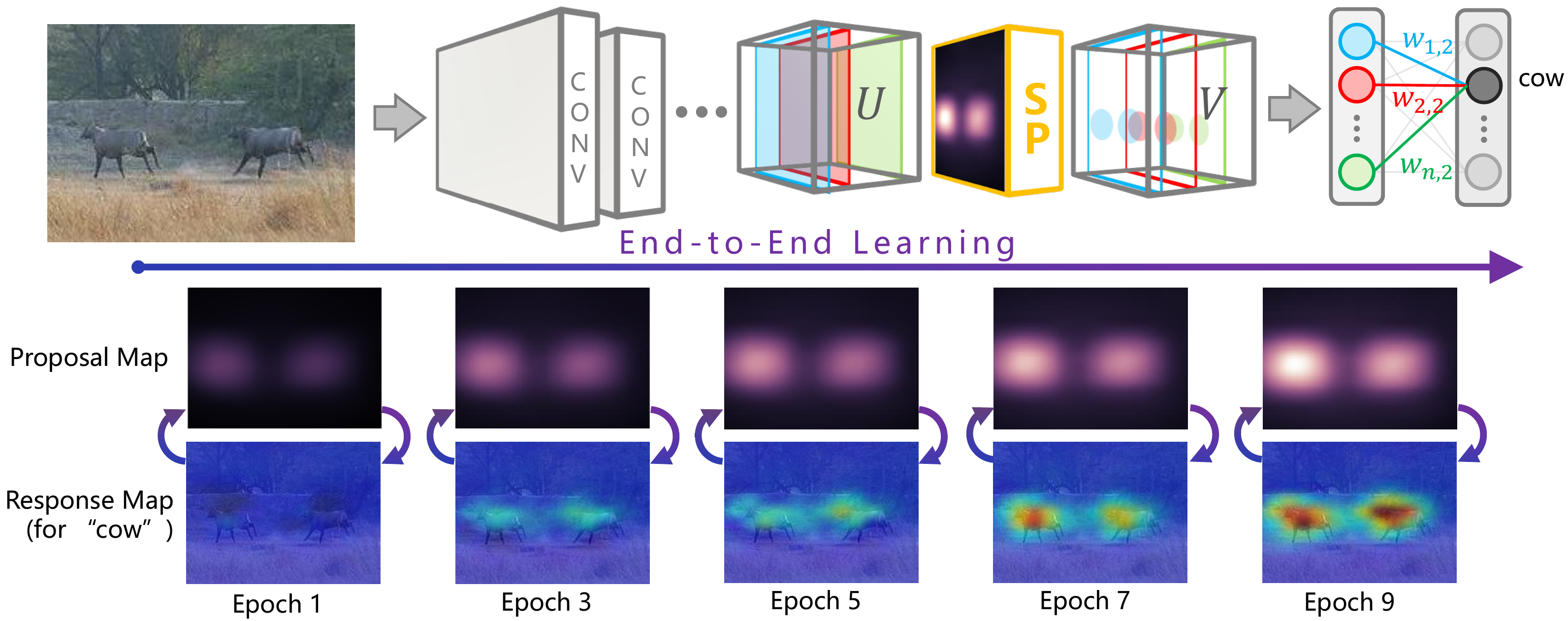}
\end{center}
  \caption{The first row shows the Soft Proposal Network architecture. The second row illustrates the evolution of the proposal map during training epochs (corresponding to the outer loop of Algorithm~\ref{alg:A1}). The third row presents the evolution of the response map for ``cow''. The proposal map produced by SP module iteratively evolves and jointly optimizes with convolutional filters during the learning phase, leading SPN to discover fine-detailed visual evidence for localization. Best viewed in color.}
\label{fig:arch}
\vspace{-1em}
\end{figure*}

\subsection{Weakly Supervised Localization}
\textbf{Pipelined methods.} 
Weakly supervised localization methods often use a stepwise strategy, \ie, first extracting candidate proposals and then learning classification model together with selecting proposals to localize objects. Many approaches have been explored to prevent the learning procedure from getting stuck to a local minimum, \eg, prior regularization \cite{bilen2014weakly}, multi-fold learning \cite{cinbis2017weakly}, and smooth optimization methods \cite{song2014learning, bilen2014weakly}.
One representative method is WSDDN \cite{bilen2016weakly}, which significantly improves the object detection performance by performing proposal selection together with classifier learning. ContextLoc \cite{kantorov2016contextlocnet} updates WSDDN by introducing two context-aware modules which try to expand or contract the fixed proposals in learning procedure to leverage the surrounding context to improve localization. 
Attention net \cite{teh2016attention} computes an attention score for each pre-computed object proposals. 
ProNet \cite{sun2016pronet} uses parallel CNN streams for multiple scales to propose possible object regions and then classify these regions via cascaded CNNs. 

To the best of our knowledge, we are the first to integrate proposal step into CNNs and achieve jointly updating among proposal generation, object region selection, and object detector estimation under weak supervision.

\textbf{Unified frameworks.} 
Another line of research shows up in weakly supervised localization uses unified network frameworks to perform both localization and classification. 
The essence of the method Oquab \etal \cite{oquab2015object} is that the deep feature maps are interpreted as a ``bag'' of instances, where only the highest responses of feature maps contribute to image label prediction in an MIL-like learning procedure. 
Zhou \etal \cite{zhou2015cnnlocalization} achieve remarkable localization performance by leveraging a global average pooling layer behind the top convolutional layer to aggregate class-specific activation. 
In the following works, Zhang \etal \cite{zhang2016EB} formulate such a class activation procedure as conditional probability backward propagation along convolutional layers to localize discriminative patterns in generic CNNs. Bency \etal \cite{bency2016weakly} propose a heuristic search strategy to hypothesize locations of feature maps in a multi-scale manner and grade the corresponding receptive fields by the classification layer. 

The main idea of these methods is that the convolutional filters can behave as detectors to activate locations on the deep feature maps, which provide informative evidence for image classification. Despite the simplicity and efficiency of these networks, they are observed missing useful object evidence, as well as being misled by complex backgrounds. The reason behind this phenomenon can be that the filters learned for common object classes are challenged with object appearance variations and background complexity. Our proposed SPN targets at solving such problems by utilizing image-specific objectness prior and coupling it with the network learning.

\section{Soft Proposal Network}
In this section, we present a network component, Soft Proposal (SP), to be plugged into standard convolutional architectures for nearly cost-free object proposal. CNNs using SP module are referred to as Soft Proposal Networks (SPNs), Fig.~\ref{fig:arch}.
Despite the SP module can be inserted after any CNN layer, we apply it after the last convolutional layer where the deep features are most informative. 
For weakly supervised object localization, SPN has an spatial pooling layer with the output features connected to image labels, as illustrated later.

In the learning procedure of SPN, the Soft Proposal Generation step spotlights potential object locations via performing graph propagation over the receptive fields of deep responses, and the Soft Proposal Coupling step aggregates feature maps with the generated proposal map. With iterative proposal generation, coupling, and activation, SPN performs weakly supervised learning in an end-to-end manner.

\subsection{Soft Proposal Generation}
\label{subsec:generate}
\begin{figure}[t!]
\begin{center}
\includegraphics[width=0.93\linewidth]{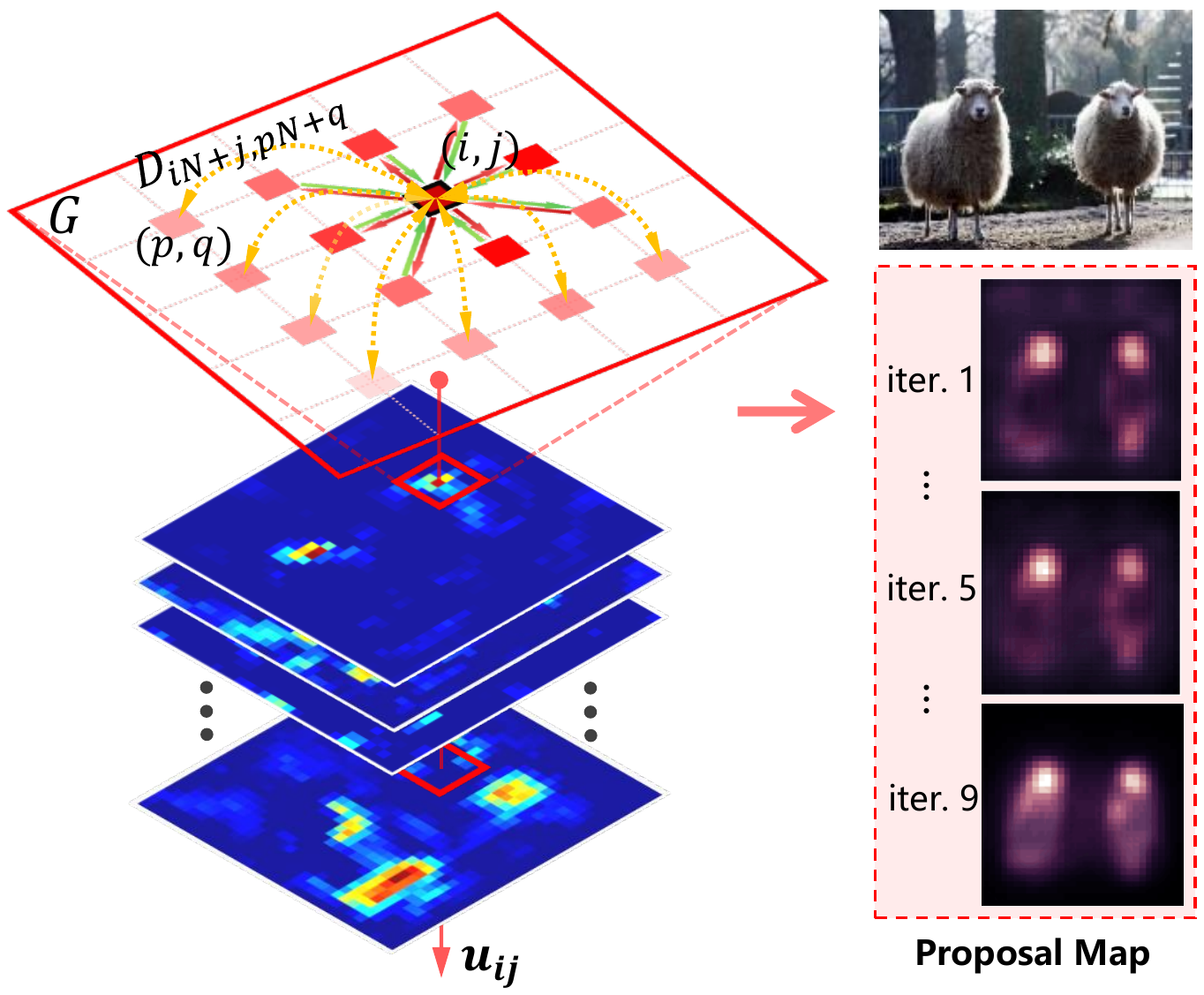}
\end{center}
  \caption{Soft Proposal Generation in a single SPN feedforward pass (corresponding to the inner loop of Algorithm~\ref{alg:A1}). Experimentally, the generation reaches stable in about ten iterations. }
\label{fig:spg}
\vspace{-0.8em}
\end{figure}

The proposal map, $M \in \mathbb R^{N \times N}$, is an objectness map generated by SP module based on the deep feature maps, Fig.~\ref{fig:spg}. Consider a SP module is inserted after the $l$-th convolutional layer, let $U^l \in \mathbb R^{K \times N \times N}$ denote the deep feature maps of the $l$-th convolutional layer, where $K$ is the number of feature maps (channels), $N \times N$ denotes the spatial size of a feature map. Each location $(i,j)$ on $U^l$ has a deep feature vector $\mathbf u^l_{ij} = U^l_{\cdot,i,j} \in \mathbb R^K$ from all $K$ channels of $U^l$. 
To generate $M$, a fully connected directed graph $G$ is first constructed by connecting every location on $U^l$, with the weight matrix $D \in \mathbb R^{N^2 \times N^2}$ where $D_{iN+j, pN+q}$ indicating the weight of edge from node $(i,j)$ to node $(p,q)$.
 
To calculate the weight matrix $D$, two kinds of objectness measures are utilized: 1). Image regions from the same object category share similar deep features. 2). Neighboring regions exhibit semantic relevance. 
The objectness confidence are reflected with a dissimilarity measure that combines feature difference and spatial distance, as 
$D^{\prime}_{iN+j, pN+q} \triangleq \Vert \mathbf u^l_{ij}-\mathbf u^l_{pq} \Vert \cdot L(i-p,j-q)$, and $L(a,b) \triangleq exp(-\frac{a^2 + b^2}{2\epsilon^2})$, where $\epsilon$ is empirically set as $0.15N$ in all experiments. And then the weights of the outbound edges of each node are normalized to 1, \ie, 
$D_{a,b} = \frac{D^{\prime}_{a,b}}{\sum_{a=1}^N D^{\prime}_{a,b}}$.

With the weight matrix $D$ defining the edge weight between nodes, a graph propagation algorithm, \ie, random walk \cite{lovasz1993random}, is utilized to generate the proposal map $M$. The random walk algorithm iteratively accumulates objectness confidence at the nodes that have high dissimilarity with their surroundings. 
A node receives confidence from inbound directed edges, and then the confidence among the nodes can be diffused along the outbound directed edges which are connected to all other nodes, Fig.~\ref{fig:spg}. In this procedure, a location transfer confidence to others via globally objectness flow, which not only collects local object evidence but also depresses noise regions.
For the convenience of random walk operation, we first reshape the 2D proposal map $M$ to a vector with $N^2$ element, initialized with the value $\frac{1}{N^2}$. $M$ is updated with iteratively multiplying with the weight matrix $D$, as
\begin{equation}
    M \gets D \times M.
\label{equ:generation}
\end{equation}
The above procedure is a variant of the eigenvector centrality measure \cite{newman2008mathematics}, which outputs a proposal map to indicate the objectness confidence of each location on the deep feature maps. Note that the weight matrix $D$ is conditional on the deep feature maps $U^l$, and $U^l$ is conditional on the convolutional filters of the $l$-th layer, $W^l$, in the learning procedure. To show such dependency, Eq.~\ref{equ:generation} is updated as
\begin{equation}
    M \gets D\big(U^l(W^l)\big)\times M.
\label{equ:generation2}
\end{equation}
The random walk procedure can be seen as a Markov chain that can reach unique stable state because the chain is ergodic, a property which emerges from the fact that the graph $G$ is by construction strongly connected \cite{harel2006graph}.
Given deep feature maps $U$, Eq.~\ref{equ:generation2} usually reaches its stable state in about ten iterations, and the output $M$ is reshaped from a vector to a 2D proposal map $M \in \mathbb R^{N \times N}$. 

\subsection{Soft Proposal Coupling}
\label{subsec:coupler}
The proposal map generated with the deep feature maps in a weakly supervised manner can be regarded as a kind of objectness map, which indicates possible object regions. From the perspective of image representation, the proposal map spotlights ``regions of interest" that are informative to image classification. $M$ can be integrated into the end-to-end learning via SP module, Fig.~\ref{fig:intro-SP}, to aggregate the image-specific discriminative patterns from deep responses. 

In the forward propagation of a SP-augmented CNN, \ie, SPN, each feature map of the coupled $V \in \mathbb R^{N \times N}$ is the Hadamard product of the corresponding feature map of $U$ and $M$,
    \begin{equation}
    V_k = U^l_k(W^l) \circ M, {\scriptstyle k=1,2,...,K},
    \label{equ:forward}
    \end{equation}
where the subscript $k$ denotes the channel index and ``$\circ$'' denotes element-wise multiplication.
The coupled feature maps $V$ pass forward to predict scores $y \in \mathbb R^{C}$ of $C$ classes, and then the prediction error $E = \ell(y,t)$ of each sample comes out according to the image labels $t$. $\ell(\cdot)$ is the loss function.
In the back-propagation procedure of SPN, the gradient is apportioned by $M$, as
    \begin{equation}
        \begin{split}
        & W^l = W^l + \Delta W(M) \\
        & \Delta W(M) = - \eta \frac{\partial E}{\partial W^l} (M)
        \end{split}
    \label{equ:dynamic-lr}
    \vspace{-0.8em}
    \end{equation}
where $\eta$ is the network learning rate. $\Delta W(M)$ means that $W^l$ is conditional on $M$, as the gradients of filters $\frac{\partial E}{\partial W^l}$ are conditional on $M$, Eq.~\ref{equ:gradient-filter}.
Since $W^l$ is conditional on $M$, the SPN learns more informative image regions in each image and depresses noisy backgrounds.

Given the Soft Proposal Generation defined by Eq.~\ref{equ:generation2}, the Soft Proposal Coupling defined by Eq.~\ref{equ:forward}, and the back propagation procedure defined by Eq.~\ref{equ:dynamic-lr}, it is clear that $U^l$, $W^l$, and $M$ are conditional on each other. During training, once the convolutional filters $W^l$ changed by Eq.~\ref{equ:dynamic-lr}, $U^l$ will also change. Once $U^l$ is updated, a random walk procedure, described in Sec.~\ref{subsec:generate}, is utilized to update the proposal map $M$. The proposal map $M$ helps SPNs to progressively spotlight feature maps $U^l$ and learn discriminative filters $W^l$, thus the proposals and filters are jointly optimized in SPNs, Fig.~\ref{fig:arch}. The procedure is described in Algorithm~\ref{alg:A1}.

\setlength{\textfloatsep}{2pt}
\renewcommand{\algorithmicrequire}{\textbf{Input:}} 
\renewcommand{\algorithmicensure}{\textbf{Output:}}
\begin{algorithm}[t]
  \begin{algorithmic}[1]
    \Require{Training images with category labels}
    \Ensure{Network parameters, proposal map for each image.}
    \Repeat
        \State initial each element in $M$ with $\frac{1}{N^2}$
        \Repeat
            \State $M \gets D\big(U^l(W^l)\big)\times M$
        \Until{stable state reached}
        \State $ V = U^l(W^l) \circ M$, feed forward. 
        \State $ W^l = W^l + \Delta W(M)$, backward.
        \For {all the convolutional layers $l$}
            \State $ U^{l} = W^l*U^{l-1}$ 
        \EndFor
    \Until{Learning converges}
  \end{algorithmic}
    \caption{Learning SPN with Soft Proposal Coupling}
    \label{alg:A1}
\end{algorithm}

\subsection{Weakly Supervised Activation}
\label{subsec:activation}
The weakly supervised learning task is performed by firstly using an spatial pooling layer to aggregate deep feature maps to a feature vector, and connecting such a feature vector to image categories with a fully connect layer, Fig.~\ref{fig:arch}. Such an architecture uses weak supervision posed from the end of the network, \ie, the image category annotations, to activate potential object regions.

In the forward propagation of SPN, proposal map $M$ is generated by the SP module inserted behind the $l$-th convolutional layer. The feature maps $U^l$ is computed as 
    \begin{equation}
        U_j^l = ( \sum_{i \in S_j} U_i^{l-1} \ast W_{ij}^l + b_j^l ) \circ M,
        \label{equ:conv}
        \vspace{-0.8em}
    \end{equation}
where $S_j$ is a selection of input maps, $b_j^l$ is the additive bias, and $W_{ij}^l$ is the convolutional filters between the $i$-th input map in $U^{l-1}$ and the $j$-th output map in $U^l$.
 
In the backward propagation of SPN, the error propagates from layer $l+1$ to layer $l$ via the $\delta$, as
    \begin{equation}
        \begin{split}
            \delta^l & = \frac{\partial E}{\partial U^l} 
            = \frac{\partial E}{\partial U^{l+1}} \frac{\partial U^{l+1}}{\partial U^{l}} \\
            & = \delta^{l+1} \frac{\partial [(U^l \ast W^{l+1} + b^l) \circ M]}{\partial U^l} \\
            & = \delta^{l+1} \ast W^{l+1} \circ M,
        \end{split}
        \label{equ:grad}
        \vspace{-0.8em}
    \end{equation}
which indicates that the proposal map $M$ spotlights not only informative regions on feature maps but also worth-learning locations. Since the $M$ flows along with gradients $\delta$, inserting one SP module after the top convolutional layer can effect all CNN filters.

Once $\delta^l$ is calculated, we can immediately compute the gradients for filters as
    \begin{equation}
        \begin{split}
        \frac{\partial E}{\partial W_{ij}^l} & = \sum_{p,q} (\delta_j^l)_{pq}(\mathbf x_i^{l-1})_{pq} \\
         & = \sum_{p,q} (\delta_j^{l+1} \ast W_{j\cdot}^{l+1})_{pq} M_{pq} (\mathbf x_i^{l-1})_{pq},
        \end{split}
        \label{equ:gradient-filter}
        \vspace{-0.8em}
    \end{equation}
and compute the gradients for bias as    
    \begin{equation}
        \begin{split}
        \frac{\partial E}{\partial b_{ij}^l} & = \sum_{p,q} (\delta_j^l)_{pq}\\
        & = \sum_{p,q} (\delta_j^{l+1} \ast W_{j\cdot}^{l+1})_{pq} M_{pq},
        \end{split}
        \label{equ:gradient-bias}
        \vspace{-0.8em}
    \end{equation}
where $W_{j\cdot}^{l+1}$ denotes the filters of layer $l+1$ that are used to calculate $U^l_j$, and $(\mathbf x_i^{l-1})_{pq}$ denotes the patch centered $(p,q)$ on $U_i^{l-1}$.
With Eq.~\ref{equ:gradient-filter} and Eq.~\ref{equ:gradient-bias}, the proposal map $M$ which indicates the objectness confidence of an image combines with the gradient maps in the weakly supervised activation procedure, driving SPN to learn more useful patterns.

For weakly supervised object localization, we calculate the response map $R_c$ for the $c$-th class, similar to \cite{zhou2015cnnlocalization}, $R_c = \sum_k w_{k,c} \cdot \hat{U}_k \circ M$ where $\hat{U}_k$ is the $k$-th feature map of the last convolutional layer, $w_{k,c}$ is the weight value of the fully connected layer which connects the $c$-th output node and the $k$-th feature vector, Fig.~\ref{fig:arch}.

\section{Experiment}
We upgrade state-of-the-art CNN architectures, \eg, VGG16 and GoogLeNet, to SPNs, and evaluate them on popular benchmarks. In Sec.~\ref{Sec:exp-proposal}, we compare SPN with conventional object proposal methods, showing that it can generate high-quality proposals with negligible computational overhead. In Sec.~\ref{Sec:exp-pointing}, on a weakly supervised point-based object localization task, we demonstrate SPNs can learn better object-centric filters, which produce precise responses on class-specific objects. In Sec.~\ref{Sec:exp-boundingbox}, SPNs are further tested on a weakly supervised object bounding box localization task, validating its capability of discovering more fine-detailed visual evidence in complex cluttered scenes. In Sec.~\ref{Sec:exp-classification}, the significant improvement of classification performance on PASCAL VOC \cite{everingham2015the} (20-classes, $\unsim$10k images), MS COCO \cite{lin2014microsoft} (80-classes, $\unsim$160k images), and ImageNet \cite{russakovsky2015imagenet} (1000-classes, $\unsim$1300k images), shows the superiority of SPNs beyond weakly supervised object localization tasks\footnote{Please refer to supplementary materials for more results.}. We train SPNs using SGD with cross-entropy loss. We use a weight decay of 0.0005 with a momentum of 0.9 and set the initial learning rate to 0.01.
\subsection{Proposal Quality}
\label{Sec:exp-proposal}
\begin{figure}[t!]
    \begin{center}
        \includegraphics[width=0.98\linewidth]{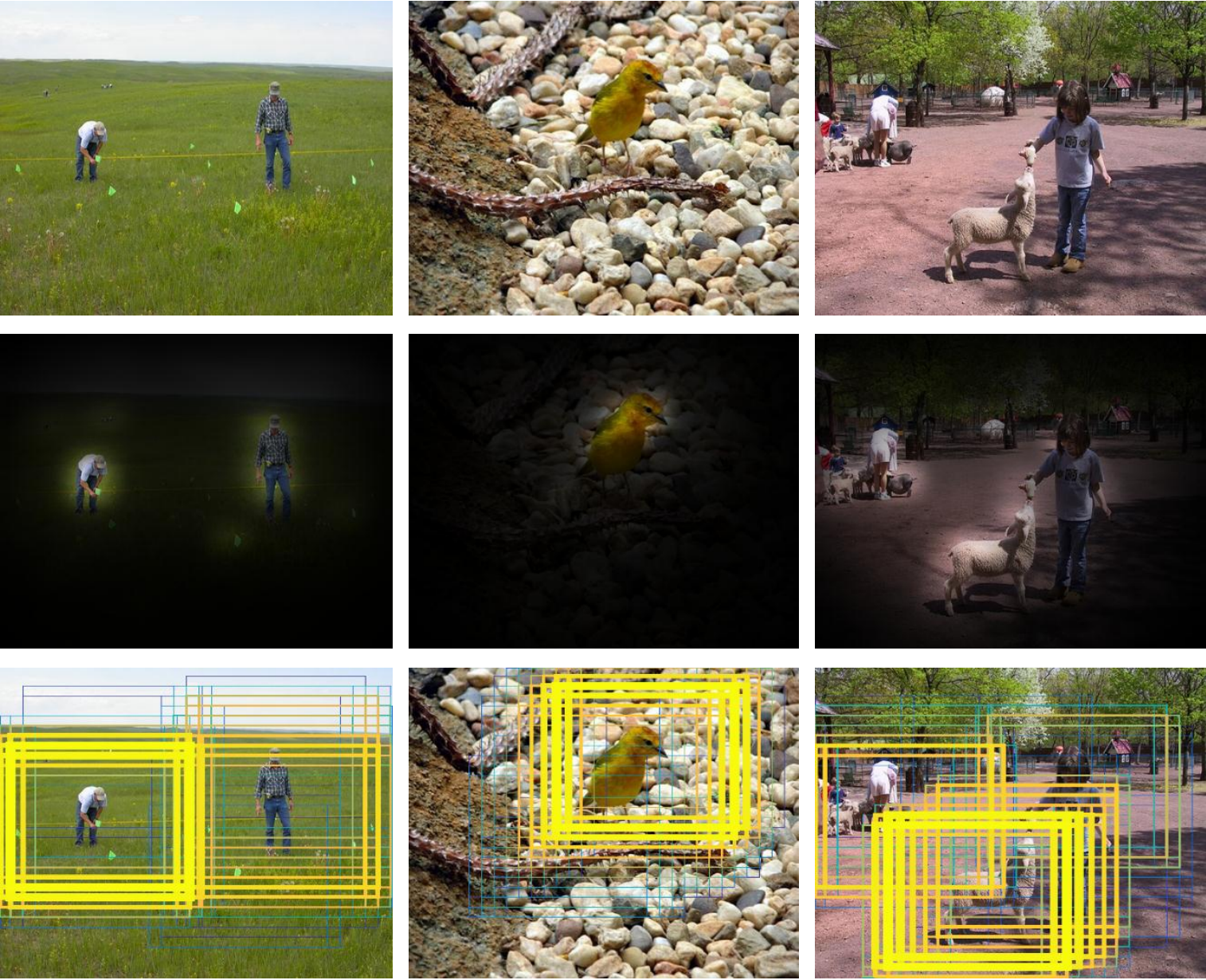}
    \end{center}
    \caption{Proposal examples. The first row presents input images. The second row presents proposal coupled images, by composing the proposal map with the original images. The third row shows top-100 scored receptive fields according to the proposal map. Best viewed in color.}
    \label{fig:sp-samples}
\end{figure}

\begin{table}
    \begin{center}
        \begin{tabular}{|l|cc|}
        \hline
        Method & ObjectEnergy(\%) & Time(ms) \\ \hline\hline
        Selective Search \cite{uijlings2013selective} & 53.7 & 2000 \\
        EdgeBoxes \cite{zitnick2014edge} & 58.8 & 200 \\
        RPN \small{(supervised)} \cite{ren2015faster} & \textbf{63.3} & 10.5 \\ \hline
        SPN \small{(weakly supervised)} & 62.2 & \textbf{0.9} \\ \hline
        \end{tabular}
    \end{center}
    \caption{
    Proposal quality evaluation on VOC2007 test set. The Object Energy in the second column indicates the percentage of spotlighted object areas. Note that RPN is learned with object bounding box annotations (supervised) while SPN is learned with image label annotations (weakly supervised). The third column describes the average time cost per image. RPN and SPN are tested with a NVIDIA Tesla K80 GPU while Selective Search and EdgeBoxes are tested on CPU due to algorithm complexity.}
    \label{tab:sp-energy}
    \vspace{-0.8em}
\end{table}

On the VOC2007 dataset, we assess the quality of proposals by an Object Energy metric defined below. For the compared Selective Search \cite{uijlings2013selective}, EdgeBoxes \cite{zitnick2014edge} and RPN \cite{ren2015faster} methods, the energy value of a pixel is the sum of scores of the proposal boxes that cover the pixel. Therefore, all objectness values in an image constitute an energy map that indicates the informative object regions predicted by the method. For the SPN, we produce Object Energy maps by rescaling proposal maps to the image size, Fig.~\ref{fig:sp-samples}. We further normalize each energy map and compute the sum of Object Energy of pixels those fall into ground-truth bounding boxes as the Object Energy. 

It can be seen from the definition that the Object Energy values range in $[0.0, 1.0]$, which indicates how many informative object areas in the image are spotlighted by the method. The second column in Tab.~\ref{tab:sp-energy} demonstrates that the proposals generated by SPN are of high-quality. The Object Energy of SPN proposals is significantly larger than those of Selective Search and EdgeBoxes, which usually produce redundant proposals and cover many background regions. Surprisingly, The Object Energy of SPN proposals obtained by weakly supervised learning is comparable to that of supervised RPN method (62.2\% vs. 63.2\%). 
\begin{figure}[t!]
    \begin{center}
        \includegraphics[width=0.98\linewidth]{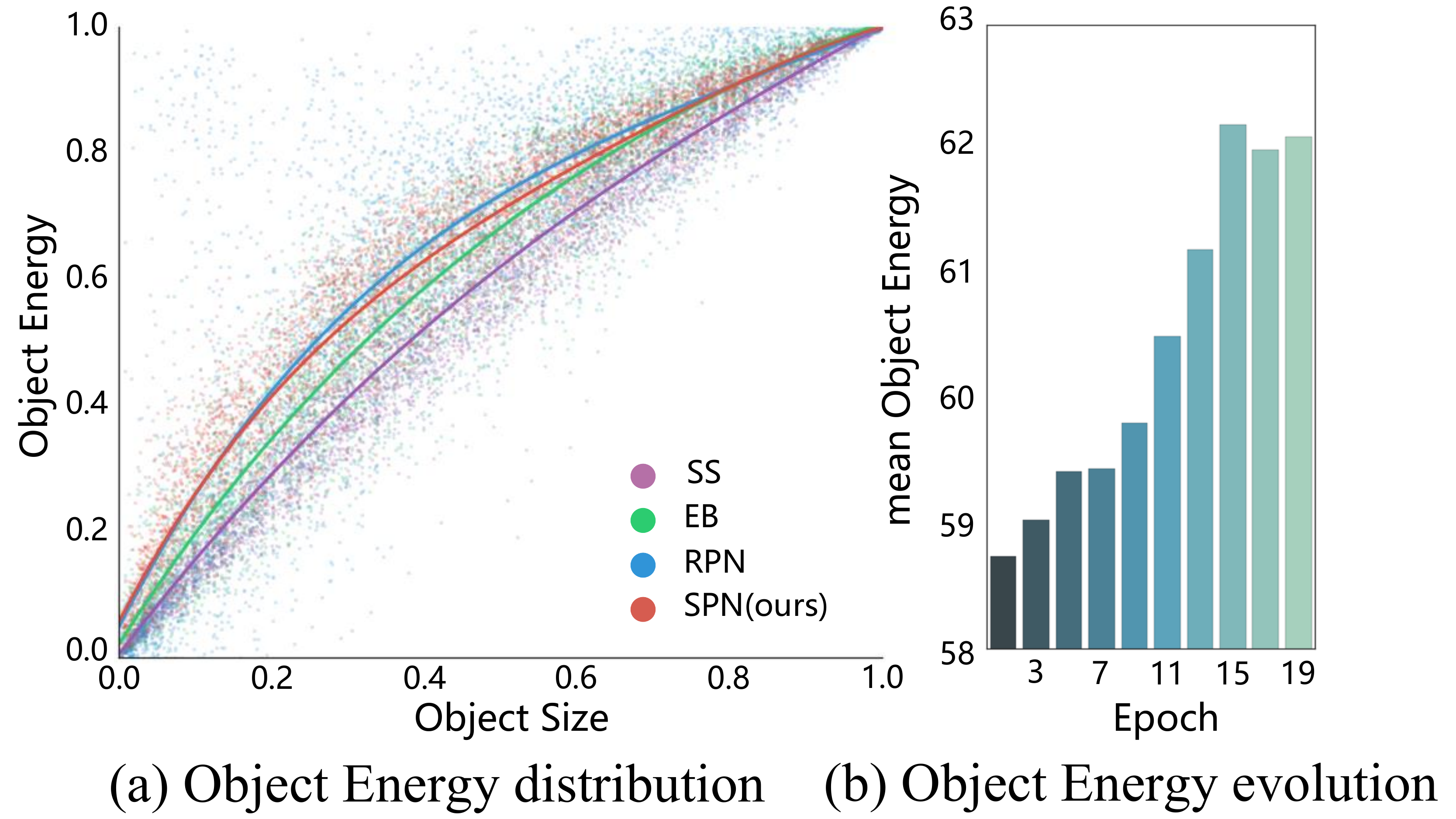}
    \end{center}
    \caption{(a) Object Energy curves. The x-coordinate is the ratio between the object area to the image size, and y-coordinate is the Object Energy. The curves are produced by using a 3-polynomial regression on the dots, each of which denotes an image. (b) Evolution of Object Energy during the learning procedure. Best viewed zooming on screen.}
    \label{fig:energy-vis}
    \vspace{-0.8em}
\end{figure}
It can be seen in Fig.~\hyperref[fig:energy-vis]{\ref*{fig:energy-vis}(a)} that the proposed SPN can spotlight small objects significantly better than the Selective Search and EdgeBoxes methods, despite that the proposal maps are based on low-resolution deep feature maps. Fig.~\hyperref[fig:energy-vis]{\ref*{fig:energy-vis}(b)} demonstrates that the SPN proposals can iteratively evolve and jointly optimize with network filters during the end-to-end training.
Moreover, the implementation of SPN is simple and naturally compatible with GPU parallelization. It can be seen from the third column of Tab.~\ref{tab:sp-energy} that the proposed SP module can introduce weakly supervised object proposal to CNNs in a nearly cost-free manner.

\subsection{Pointing Localization}
\label{Sec:exp-pointing}
    \textbf{Pointing without prediction.}
    To evaluate whether the proposed SPN can learn more discriminative filters which are effective to produce accurate response maps, we test it on the weakly supervised pointing task. We select three successful CNNs, including CNN-S \cite{Chatfield14return}, VGG16 \cite{Simonyan14very}, and GoogLeNet \cite{szegedy2015going} and upgrade them to SPNs by inserting the SP module after their last convolution layers, Fig.~\ref{fig:arch}. All SPNs are fine-tuned on the  VOC2007 training set with same hyper-parameters, and we calculate the response maps as described in Sec.~\ref{subsec:activation} with ground-truth labels for pointing localization. Following the setting of c-MWP \cite{zhang2016EB}, a state-of-the-art method, we calculate the accuracy of pointing localization as below: a hit is counted if the pixel of maximum response falls in one of the ground truth bounding boxes of the cued object category within 15 pixels tolerance. Otherwise, a miss is counted. We measure the per-class localization accuracy by $Acc=\frac{Hits}{Hits+Misses}$. The overall results are the mean value of per-class point localization accuracy.
    
    \begin{figure}[t!]
      \begin{center}
        \includegraphics[width=0.98\linewidth]{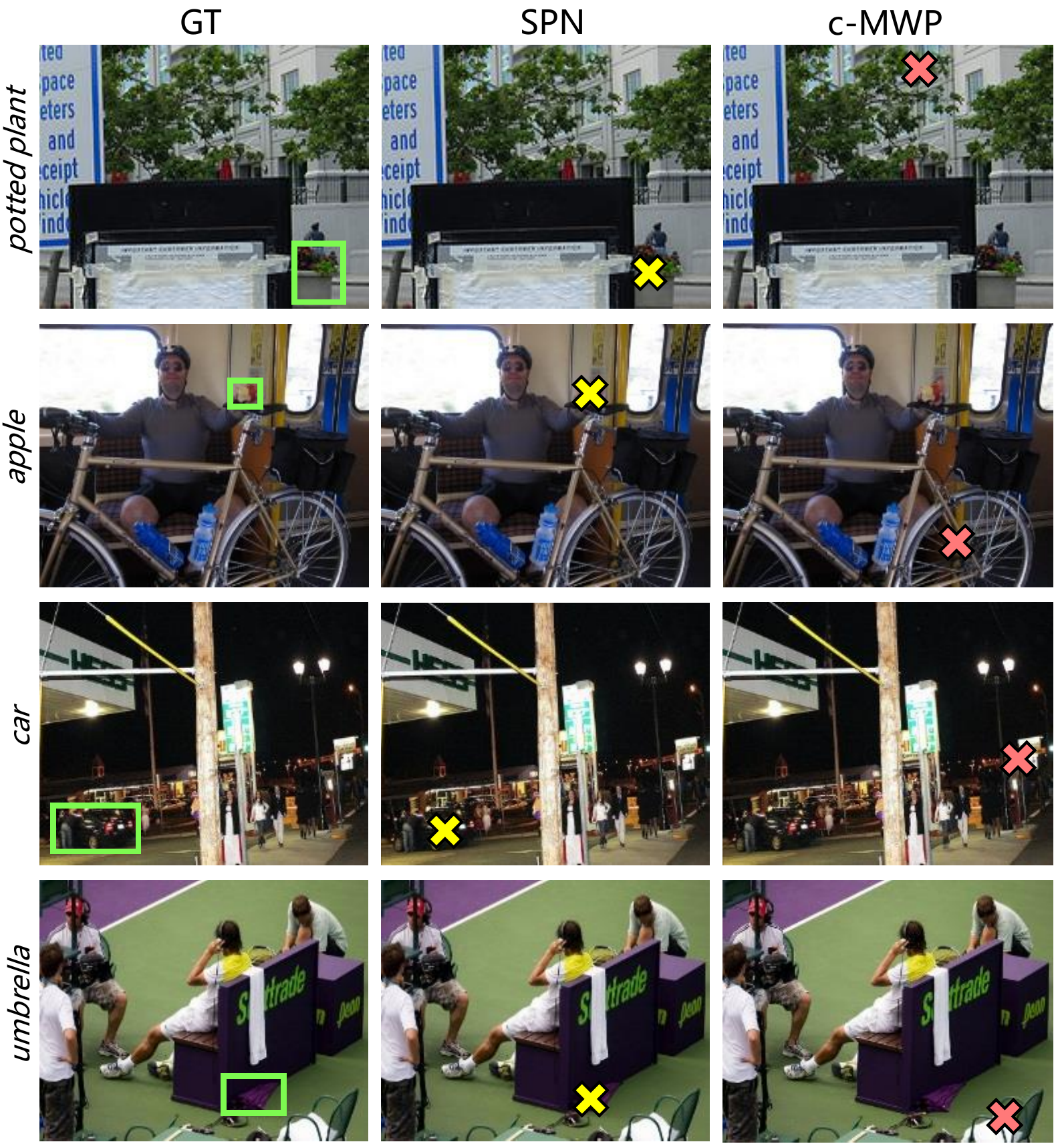}
      \end{center}
      \caption{
      Examples of pointing localization, which shows that SPN is effective in complex scenes: a) Noisy co-occurrence patterns, \eg, leaves for ``potted plant". b) Small objects, \eg, ``apple" in hand. c) Cluttered backgrounds, \eg, ``car" on the street. d) \att{Infrequent form}, \eg, closed ``umbrella". Best viewed in color.}
      \label{fig:pointing-samples}
      \vspace{-0.8em}
    \end{figure}
    
    For the VOC2007 dataset, we use two test sets, \ie, \textbf{All} and Difficult (\textbf{Diff.}) \cite{zhang2016EB}. \textbf{All} means the overall test set and \textbf{Diff.} means a difficult subset which has mixed categories and contains small objects.
    As shown in Tab.~\ref{tab:loc-acc-voc07}, upgrading conventional CNNs to SPNs brings significant performance improvement. Specifically, the SP-VGGNet outperforms c-MWP by 7.5\% (87.5 \% vs 80.0 \%) for \textbf{All} and 11.3\% (78.1\% vs 66.8\%) for \textbf{Diff.}. The SP-GoogLeNet outperforms c-MWP by 3.1\% and 6.8\% for \textbf{All} and \textbf{Diff.}, respectively. The significant improvement of pointing localization performance validates the effectiveness of the SP module for guiding SPNs to learn better object-centric filters, which can pick up accurate object responses.

    We made multiple observations in Tab.~\ref{tab:loc-acc-voc07}. 1). SP-VGGNet has better performance than SP-GoogLeNet on pointing localization. The reason can be that the receptive fields of SP-VGGNet are smaller than that of SP-GoogLeNet. Without much overlap between receptive fields, the objectness propagation in SP module can be more effective. 2). The accuracy improvement on \textbf{Diff.} is larger than that on \textbf{All}, which shows that the proposal functionality of SPNs is particularly effective in cluttered scenes.
    
    \begin{table}
      \begin{center}
        \begin{tabular}{|l|ccc|}
        \hline
        Method & CNN-S     & VGG16     & GoogLeNet \\ \hline\hline
        Center   & 69.5/42.6 & 69.5/42.6 & 69.5/42.6 \\ \hline
        Grad \cite{simonyan2013deep} & 78.6/59.8 & 76.0/56.8 & 79.3/61.4 \\
        Deconv \cite{zeiler2014visualizing} & 73.1/45.9 & 75.5/52.8 & 74.3/49.4 \\
        LRP \cite{bach2015pixel} & 68.1/41.3 & - & 72.8/50.2 \\
        CAM \cite{zhou2015cnnlocalization}   & -  & - & 80.8/61.9 \\
        MWP \cite{zhang2016EB}      & 73.7/52.9 & 76.9/55.1 & 79.3/60.4 \\
        c-MWP \cite{zhang2016EB}    & 78.7/61.7 & 80.0/66.8 & 85.1/72.3 \\ \hline
        SPN & \textbf{81.8/66.7} & \textbf{87.5/78.1} & \textbf{88.2/79.1} \\ \hline 
        \end{tabular}
      \end{center}
      \caption{Pointing localization accuracy (\%) on VOC2007 test set (\textbf{All/Diff.}). \textbf{Center} is a baseline method which uses the image centers as estimation of object centers. }
      \label{tab:loc-acc-voc07}
      \vspace{-0.8em}
    \end{table}
    
    \begin{table}[]
      \begin{center}
        \begin{tabular}{|l|cc|}
        \hline
        Method   & \multicolumn{2}{c|}{mAP (\%)} \\ \hline
        Dataset  & VOC & COCO \\ \hline\hline
        Oquab \etal \cite{oquab2015object} & 74.5 & 41.2  \\
        Sun \etal \cite{sun2016pronet} & 74.8 & 43.5  \\
        Bency \cite{bency2016weakly}  & 77.1 & 49.2   \\ \hline
        SPN & \textbf{82.9} &  \textbf{55.3} \\ \hline
        \end{tabular}
      \end{center}
      \caption{Mean Average Precision (mAP) of location prediction on VOC2012 val. set and COCO2014 val. set.}
      \label{tab:loc-map-voc12}
      \vspace{-0.8em}
    \end{table}
    
    \begin{table*}[!htp]
    \begin{center}
    \resizebox{\textwidth}{!}{%
    \begin{tabular}{c|cccccccccccccccccccc|c}
    \hline
    Method & aero & bike & bird & boat & bottle & bus & car & cat & chair & cow & table & dog & horse & mbike & persn & plant & sheep & sofa & train & tv & mean \\ \hline\hline
    Bilen \etal \cite{bilen2015weakly} & 66.4 & 59.3 & 42.7 & 20.4 & 21.3 & 63.4 & 74.3 & 59.6 & 21.1 & 58.2 & 14.0 & 38.5 & 49.5 & 60.0 & 19.8 & 39.2 & 41.7 & 30.1 & 50.2 & 44.1 & 43.7 \\
    Wang \etal \cite{wang2014weakly} & 80.1 & 63.9 & 51.5 & 14.9 & 21.0 & 55.7 & 74.2 & 43.5 & 26.2 & 53.4 & 16.3 & 56.7 & 58.3 & 69.5 & 14.1 & 38.3 & 58.8 & 47.2 & 49.1 & 60.9 & 48.5 \\
    Cinbis \etal \cite{cinbis2017weakly} & 65.3 & 55.0 & 52.4 & \textbf{48.3} & 18.2 & 66.4 & \textbf{77.8} & 35.6 & 26.5 & \textbf{67.0} & 46.9 & 48.4 & 70.5 & 69.1 & 35.2 & 35.2 & \textbf{69.6} & 43.4 & 64.6 & 43.7 & 52.0 \\
    WSDDN \cite{bilen2016weakly} & 65.1 & 58.8 & 58.5 & 33.1 & \textbf{39.8} & 68.3 & 60.2 & 59.6 & \textbf{34.8} & 64.5 & 30.5 & 43.0 & 56.8 & 82.4 & 25.5 & \textbf{41.6} & 61.5 & 55.9 & 65.9 & 63.7 & 53.5 \\
    ContextLoc \cite{kantorov2016contextlocnet} & 83.3 & \textbf{68.6} & 54.7 & 23.4 & 18.3 & \textbf{73.6} & 74.1 & 54.1 & 8.6 & 65.1 & 47.1 & 59.5 & 67.0 & \textbf{83.5}  & 35.3  & 39.9  & 67.0  & 49.7  & 63.5  & \textbf{65.2} & 55.1 \\ \hline
    \textbf{SP-VGGNet} & \textbf{85.3} & 64.2 & \textbf{67.0} & 42.0 & 16.4 & 71.0 & 64.7 & \textbf{88.7} & 20.7 & 63.8 & \textbf{58.0} & \textbf{84.1} & \textbf{84.7} & 80.0 & \textbf{60.0} & 29.4 & 56.3 & \textbf{68.1} & \textbf{77.4} & 30.5 & \textbf{60.6} \\ \hline
    \end{tabular}%
    }
    \end{center}
    \caption{Correct Localization rate (CorLoc \cite{deselaers2012weakly}) on the positive trainval images of the VOC2007 dataset (\%).}
    \label{tab:corloc-voc07}
    \vspace{-0.8em}
    \end{table*}
    
    \textbf{Pointing with prediction.} 
    We further test SPN on a more challenging pointing-with-prediction task. The task requires the network output not only the correct prediction of the presence/absence of the object categories in test images, but also the correct pointing localization of objects, \ie, the point of maximum response falls in one of the ground truth bounding boxes within 18 pixels tolerance \cite{oquab2015object}.
   
    We upgrade a pre-trained VGG16 model to SPN and respectively fine-tune it on VOC2012 and COCO2014 dataset for 20 epochs. Results are reported in Tab.~\ref{tab:loc-map-voc12}. 
    Without multi-scale setting, SPN outperforms the state-of-the-art method \cite{bency2016weakly} by a significant margin (5.8\% mAP for VOC2012, 6\% mAP for COCO2014).
    This evaluation demonstrates that the Soft Proposal module endows CNNs accurate localization capability while keeping its classification ability. In Sec.~\ref{Sec:exp-classification}, we will show that upgrading CNNs to SPNs can even improve the classification performance.

\subsection{Bounding Box Localization}
    Although without object-level annotations involved in the learning phase, our method can also be used to estimate object bounding boxes with the help of response maps. We calculate each response map with ground truth labels and convert them to binary maps with the mean value as thresholds. We then rescale them to the original image size and extract the tightest box covering the foreground pixels as the predicted object bounding box.
    
\begin{figure}[t!]
    \begin{center}
        \includegraphics[width=0.98\linewidth]{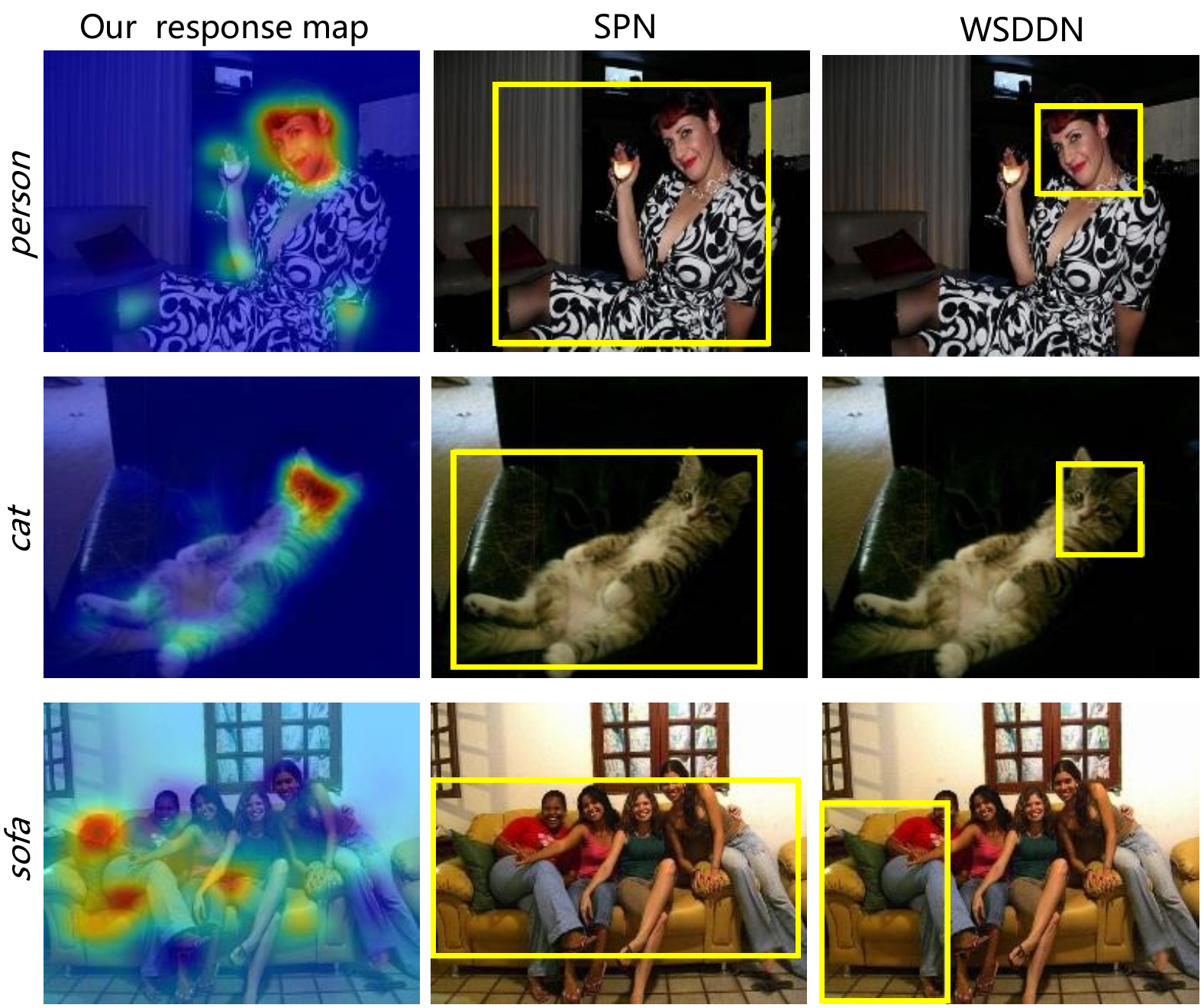}
    \end{center}
      \caption{Bounding box localization results on the VOC2007 test set. By activating fine-detailed evidence like arm or leg for ``person'', paw for ``cat'', and texture fragments for ``sofa'', the estimated bounding boxes are more precise than those by WSDDN.}
    \label{fig:map-box}
    \vspace{-0.8em}
    \end{figure}

    The Correct Localization (CorLoc) metric \cite{deselaers2012weakly} is used to evaluate the bounding box localization performance. 
    It can be seen in Tab.~\ref{tab:corloc-voc07} that the mean CorLoc of our method outperforms the state-of-the-art ContextLoc method \cite{kantorov2016contextlocnet} by about 5\%. Surprisingly, on the ``dog", ``cat", ``horse", and ``person" classes, SPN outperforms the compared method up to 20-30\%. It can be seen from Fig.~\ref{fig:map-box} that the conventional method tends to use the most discriminative part for each category, \eg, faces, while SPN can discover more fine-detailed object evidence, \eg, hands and legs, thanks to the objectness prior introduced by the SP module. On the "sofa" and "table" classes, our method outperforms other methods by 10\%, demonstrating the capability of SPN to correctly localize the occluded objects, Fig.~\ref{fig:map-box}, which shows that the graph propagation in the Soft Proposal Generation step helps to find object fragments of similar appearance. 
    
    \label{Sec:exp-boundingbox}
    
\subsection{Image Classification}
\label{Sec:exp-classification}
    Although to predict the presence/absence of object categories in an image does not require accurate located and comprehensive visual cues, the proposal functionality of SPNs which highlights informative regions while suppressing disturbing backgrounds during training should also benefit the classification performance.
    
    We use GoogLeNetGAP \cite{zhou2015cnnlocalization}, a simplified version of GoogLeNet, as the baseline. By inserting SP module after the last convolution layer, the GoogLeNetGAP is upgraded to a SPN. The SPN is trained on the ILSVRC2014 dataset, \ie, ImageNet, for 90 epochs with the SGD method. 
    It can be seen in the second column of Tab.~\ref{tab:cls-err-imagenet} that the SPN significantly outperforms the baseline GoogLeNetGAP by 1.5\%, which shows that the SPNs can learn more informative feature representation. We then fine-tune each trained model on COCO2014 and VOC2007 by 50 and 20 epochs to assess the generalization capability of SPN. As shown in the third column of Tab.~\ref{tab:cls-err-imagenet}. SP-GoogLeNetGAP surpasses the baseline by a large margin, \eg, 4.5\% on VOC2007. This further demonstrates that the weakly supervised object proposal is effective for both localization and classification.
    \begin{table}
        \begin{center}
            \begin{tabular}{|l||cccc|c|}
            \hline
            Method    & CAM  & c-MWP & MWP & Fb\cite{zhang2016EB} & SPN \\ \hline
            Error (\%) & 48.1&  57.0 & 38.7& 38.8 & \textbf{36.3}  \\ \hline
            \end{tabular}
        \end{center}
        \caption{Bounding box localization errors on ILSVRC2014 val. set.}
        \label{tab:loc-err-imagenet}
    \end{table}
    
    \begin{table}[]
        \begin{center}
            \begin{tabular}{|l|c|cc|}
            \hline
            Method       & ImageNet & COCO & VOC  \\ \hline\hline
            GoogLeNetGAP\cite{zhou2015cnnlocalization} & 35.0/13.2 & 54.4  & 83.4 \\ 
            SP-GoogLeNetGAP  & \textbf{33.5/12.7} & \textbf{56.0} & \textbf{84.2} \\ \hline
            \end{tabular}
        \end{center}
        \caption{Classification results. The second column is the top-1/top-5 error rate (\%) on ILSVRC2014 val. set. The third and fourth column are mAP (\%) on VOC2007 test set and COCO val. set.}
        \label{tab:cls-err-imagenet}
        \vspace{-0.8em}
    \end{table}

\section{Conclusions}
In this paper, we proposed a simple yet effective technique, Soft Proposal (SP), to integrate nearly cost-free object proposal into CNNs for weakly supervised object localization. We designed the SP module to upgrade conventional CNNs, \eg, VGG and GoogLeNet, to Soft Proposal Networks (SPNs).
In SPNs, iteratively evolved object proposals are generated based on the deep feature maps then projected back, leading filters to discover more fine-detailed evidence through the unified learning procedure.
SPNs significantly outperforms state-of-the-art methods on weakly supervised localization and classification tasks, demonstrating the effectiveness of coupling object proposal with network learning.

\section*{Acknowledgements}
The authors are very grateful for support by NSFC grant 61671427, BMSTC grant Z161100001616005.

{\small
\bibliographystyle{ieee}
\bibliography{egbib}
}
\end{document}